\documentclass{article}
\usepackage{amssymb}
\usepackage{microtype}
\usepackage{graphicx}
\usepackage{caption}
\usepackage{amsmath}
\usepackage{amssymb}
\usepackage{mathtools}
\usepackage{amsthm}
\usepackage{amsfonts}
\usepackage{nicefrac}
\usepackage{float}
\usepackage{booktabs}
\usepackage[dvipsnames]{xcolor}
\usepackage{colortbl}
\usepackage{algorithm}
\usepackage[noend]{algorithmic}
\usepackage{bbding}
\usepackage{fancyvrb}
\usepackage{listings}
\usepackage{subcaption}
\usepackage{wrapfig}
\usepackage{booktabs}
\usepackage{graphicx}
\usepackage{array}
\usepackage{float}
\usepackage{wrapfig}
\usepackage{hyperref}
\newcommand{\dataset}[1]{\mathcal{D}_{\mathit{#1}}}

\usepackage{multirow}

\newcommand{\methodname}[1]
{\textsc{Collage}}

\newcommand{\numtasks}[0]{\textbf{$6$ }}

 \usepackage[preprint]{corl_2025} 

\title{\methodname{}: Adaptive Fusion-based Retrieval \\ for Augmented Policy Learning}

\author{
Sateesh Kumar, Shivin Dass, Georgios Pavlakos\footnotemark[1]~ , Roberto Martín-Martín\footnotemark[1] \\
The University of Texas at Austin \\
{\tt\small \{sateesh, sdass, pavlakos, robertomm\}@utexas.edu}
}

\begin{document}
\maketitle


\begin{abstract}
In this work, we study the problem of data retrieval for few-shot imitation learning: select data from a large dataset to train a performant policy for a specific task, given only a few target demonstrations.
Prior methods retrieve data using a single-feature distance heuristic, assuming the best demonstrations are those that most closely resemble the target examples in visual, semantic, or motion space.
However, this approach captures only part of the relevant information and often retrieves harmful demonstrations, such as unrelated tasks with similar layouts or motions from tasks with divergent goals. We present \methodname{}, a method for \textsc{coll}ective data \textsc{ag}gr\textsc{e}gation in few-shot imitation learning that uses an adaptive late fusion mechanism to guide the selection of relevant demonstrations based on a task-specific combination of multiple cues. 
\methodname{} follows a simple, but flexible and efficient data aggregation recipe: it assigns weights to subsets of the dataset that are pre-selected using a single feature (e.g., appearance, shape, or language similarity), based on their task relevance, measured by how well a policy trained on each subset predicts actions in the few target demonstrations. 
These weights are then used during policy training to perform importance sampling over the aggregated dataset,
sampling data more densely or sparsely,
according to their estimated relevance.
This weighted aggregation strategy is general and feature-agnostic, allowing \methodname{} to combine and leverage any number of subsets selected by any retrieval heuristic or method, and to identify which subset provides the most benefit for the target task.
In extensive experiments, \methodname{} outperforms state-of-the-art retrieval and multi-task learning approaches, achieving an $11.2\%$ relative improvement over the best baseline in simulation across 10 tasks, and a $57.4\%$ relative improvement in the real world across 6 tasks.
For our real-world experiments, we include data selection from the large-scale DROID dataset, significantly improving few-shot imitation policy training. For more information: \url{https://robin-lab.cs.utexas.edu/COLLAGE}.
\end{abstract}

\renewcommand{\thefootnote}{$*$ }
\footnotetext[1]{Equal Advising. Correspondence to: \url{sateesh@utexas.edu}}
\renewcommand{\thefootnote}{\arabic{footnote}}

\renewcommand{\thefootnote}{\arabic{footnote}} 
\section{Introduction}
\label{s:intro}
\begin{figure}[!htbp]
    \centering
    \includegraphics[ width=\textwidth]{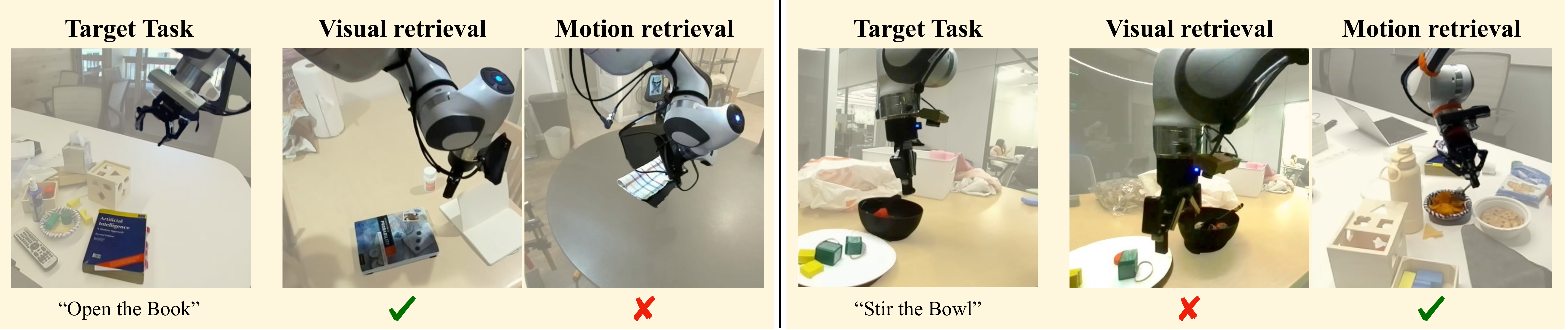}
    \caption{\textbf{Different tasks benefit from retrieval based on different modalities.}
    For the target task ``Open the book'' (left), retrieval based on visual similarity tends to return relevant demonstrations where a robot opens a book, whereas retrieval based on motion similarity often returns demonstrations with similar motions but different semantics, such as unfolding a cloth.
    On the other hand, for the target task ``Stir the bowl'' (right), retrieval based on motion similarity is more effective, returning demonstrations that involve stirring motions, while visual similarity tends to retrieve examples that feature a bowl but involve different actions, such as placing an object inside the bowl.
    }
    \label{fig:teaser}
\vspace{-2em}
\end{figure}

Imitation Learning (IL)~\citep{schaal2003computational, argall2009survey, zhang2018deep} has emerged as a powerful framework for training robot policies by mimicking expert demonstrations. 
With the rise of transformer- and diffusion-based models, single task IL approaches have achieved impressive performance on complex robotic manipulation domains~\citep{zhao2024aloha,chi2023diffusion, fu2024mobile}.
However, these successes often depend on the availability of extensive demonstration data on the specific task of interest. 
In real-world environments, such as homes, where object configurations and task requirements frequently change, collecting sufficient demonstrations for every possible task and variation is costly, time-consuming, and ultimately impractical.

As a solution to overcome the need for task-specific demonstrations, the community has explored training generalist multi-task policies~\citep{reed2022generalist, brohan2022rt, black2024pi_0} on large-scale datasets~\cite{khazatsky2024droid,o2024open}, which cover hundreds of tasks and environments and even different robot morphologies. 
The hope is that policies trained on a diverse array of experiences will generalize to unseen tasks with minimal adaptation.  
However, in practice, generalist policies often underperform on specific tasks when compared to expert policies trained on a sufficiently large number of demonstrations. This seems to be largely caused by a \textit{negative transfer} effect~\cite{wang2019characterizing}, where irrelevant or conflicting demonstrations degrade the policy’s ability to focus on task-relevant behavior.

To address this issue, a recently explored alternative for task-specific imitation policy learning leverages the same existing large demonstration datasets in a novel way: to retrieve data and augment a small dataset with a few target-task-specific demonstrations~\citep{du2023behavior, lin2024flowretrieval, memmel2024strap, nasiriany2022learning}. 
Prior approaches in retrieval-augmented few-shot imitation learning rely on different types of similarity to retrieve relevant trajectories, including visual features~\citep{du2023behavior, memmel2024strap}, optical flow~\citep{lin2024flowretrieval}, or language embeddings~\citep{zha2024distilling}, i.e., they
assume that the most helpful trajectories to train for a task are those that look, move, or are called similarly in the large dataset.
However, while each of these assumptions is true in a broad, statistical sense, they can easily fail for specific cases (see Fig.~\ref{fig:teaser}), rendering methods relying on only one of these single-modality heuristics for retrieval highly variable and brittle.
Motivated by the limitations of single-modality retrieval, we ask: \emph{Can we design a strategy to automatically combine data retrieved using different similarity measures for higher performance in retrieval-augmented few-shot policy learning?}

We present \methodname{}, a method that demonstrates that the above is indeed possible by \textsc{coll}ectively \textsc{ag}gr\textsc{e}gating subsets of the dataset preselected using single-modality heuristics in a synergistic manner.
For aggregation, \methodname{} proposes associating weights with each single-modality subset based on an estimate of their task-training relevance.
To compute this estimation, \methodname{} employs a rollout-free mechanism: it trains reference policies on each retrieval subset and estimates the policy task-relevance by evaluating the log-likelihood of the few target demonstrations.
We can consider conceptually these weights to be an approximation of the probability of a subset to be helpful to train a performant policy for the given task, and use them to guide an importance sampling mechanism during policy training: more relevant subsets of data are sampled more densely compared to less relevant subsets during the learning process.
As a result, \methodname{} is agnostic to the type of retrieval features, enabling flexible integration of different similarity modalities. 

We evaluate \methodname{} in both simulated and real-world settings. In simulation, we use the LIBERO benchmark~\citep{liu2023libero} and show that our approach outperforms both single-feature retrieval and generalist policy learning baselines, achieving relative gains of $11.2\%$ and $41.1\%$, respectively.

For our real-world evaluation, we retrieve from the DROID dataset~\citep{khazatsky2024droid} and demonstrate that our method achieves a $57.4\%$ relative improvement over state-of-the-art in few-shot imitation learning performance, achieving robust retrieval directly from large, diverse offline datasets without requiring manual curation, even in the presence of substantial visual and task domain shifts.
\section{Related Work}
\methodname{} is a novel methodology for collective data aggregation in few-shot imitation learning settings. We begin by reviewing prior work in this area, followed by related work in zero-shot manipulation and multi-modal representation learning.

\textbf{Data retrieval for few-shot imitation learning} methods assume access to a small number of target demonstrations that are augmented with data from a large dataset to improve policy training performance~\citep{nasiriany2022learning, du2023behavior, lin2024flowretrieval, memmel2024strap, zha2024distilling, wang2023voyager, yin2024offline, zhu2024retrieval, guo2025srsa}.
These methods employ different similarity measures to select the subset of data most similar to the target few demonstrations from a large-scale robotics dataset such as DROID~\cite{khazatsky2024droid} or the Open X-Embodiment (OXE) dataset~\cite{o2024open}. For instance, \citet{du2023behavior} use state-action similarity, while \citet{lin2024flowretrieval} use optical flow to measure motion similarity. \citet{zha2024distilling} and \citet{wang2023voyager} retrieve based on language similarity. Closer to our work, STRAP~\citep{memmel2024strap} retrieves sub-trajectories using DINO-based visual similarity~\cite{oquab2023dinov2}. \methodname{} generalizes this by performing retrieval across multiple modalities and adaptively weighting their contributions during training. During the preparation of this manuscript, a concurrent work~\cite{dass2025datamil} appeared that estimates an optimal demonstration subset to retrieve using a supervised, data-driven objective inspired by datamodels~\cite{ilyas2022datamodels}. In contrast, \methodname{} fuses multiple subsets, each potentially retrieved using different strategies including~\cite{dass2025datamil}, by assigning weights based on their utility for the target task

\textbf{Retrieval for zero-shot Manipulation.}
A related body of work in robotics applies similar ideas to a zero-shot setting~\citep{izquierdo2022conditional, malato2024zero, di2024dinobot, kuang2024ram, papagiannis2024r+}.
Using different measures of similarity {with respect to the current observations}, they are able to retrieve data from large datasets to develop useful manipulation policies.
For example, DINO-bot~\citep{di2024dinobot} uses DINO features to identify the most similar demonstration in the dataset and replays its actions for task completion. 
Similarly,~\citet{papagiannis2024r+} retrieve demonstrations by first applying language-based filtering, followed by visual similarity, and then track $3D$ keypoints from the selected demonstration to guide execution.
\citet{kuang2024ram} measure feature similarity from a stable diffusion model~\citep{rombach2022high} to replicate affordances from human videos, eliminating the need for a robotic dataset. 
While these methods do not require in-domain demonstrations, they assume that the large prior dataset of demonstrations contains trajectories that can be directly replayed to solve the target task, a strong assumption in unstructured, highly variable environments like homes.

\textbf{Multi-modal representations} have been widely explored in artificial perception~\cite{barnum2020benefits,spinello2012leveraging,mo2024unveiling,lin2024moma,nagrani2021attention} and robotics~\cite{shah2023mutex,kuang2024ram}. Fusion methods generally follow either early or late fusion~\cite{sanchez2016comparative}. In early fusion, raw modality data is combined before processing~\cite{barnum2020benefits,spinello2012leveraging,mo2024unveiling}, while in late fusion, each modality is processed independently and then combined—commonly seen in transformer architectures that fuse modality-specific tokens~\cite{lin2024moma,nagrani2021attention}. \methodname{} follows a \textit{late fusion} paradigm: each modality is used independently to retrieve a subset of relevant data, and these subsets are then collectively aggregated (fused) synergistically based on their relevance. In the context of data retrieval, \citet{yu2023devil} propose a multi-stage pipeline that filters data using features like DINO and Stable Diffusion in a fixed sequence, while \citet{kuang2024ram} present a modality-specific retrieval process conditioned on language, vision, and proprioception. These methods often require manually tuning the order and thresholds for each feature, limiting scalability. By contrast, \methodname{} provides a data-driven, feature-agnostic approach that easily scales to any number or type of modalities. Its fusion mechanism assigns relevance scores to each modality’s retrieved subset, allowing the final result to reflect their combined utility. This weighting strategy draws from importance-based data selection, where high-utility data points—such as human interventions~\cite{mandlekar2020human} or rare classes~\cite{chen2023area}—are emphasized. More broadly, it connects to importance sampling~\cite{tokdar2010importance}, where samples are reweighted to approximate expectations under a target distribution, enabling more effective learning from heterogeneous and imbalanced data.
\section{\methodname{}}
\label{sec:approach}

\textbf{Problem Formulation.} We consider a few-shot imitation learning setting where the goal is to learn a policy given a small set of expert demonstrations. 
Specifically, we are given a target dataset, $\dataset{target} = \{t_1, t_2, \cdots, t_n\}$, consisting of $n$ expert trajectories. Each trajectory $t_i$ is a sequence of state-action pairs, $\{(s_1, a_1), (s_2, a_2), \cdots, (s_h, a_h)\}$, accompanied by a task instruction $\ell$. Considering the limited size of $\dataset{target}$, it is not possible to train a high-performing policy on this dataset alone.
Instead, we are also given access to a large-scale offline dataset, $\dataset{prior} = \{t^p_1, t^p_2, \cdots, t^p_m\}$, with $m \gg n$, that can support augmented policy learning.
The assumption is that a policy, $\pi_\mathit{aug}$, trained via imitation learning on an augmented dataset combining $\dataset{target}$ with the right subset of $\dataset{prior}$ would have higher performance than training exclusively in the few demonstrations of $\dataset{prior}$, enabling efficient few-shot imitation learning.

\begin{figure}[!t]
    \centering
    \includegraphics[ width=\textwidth]{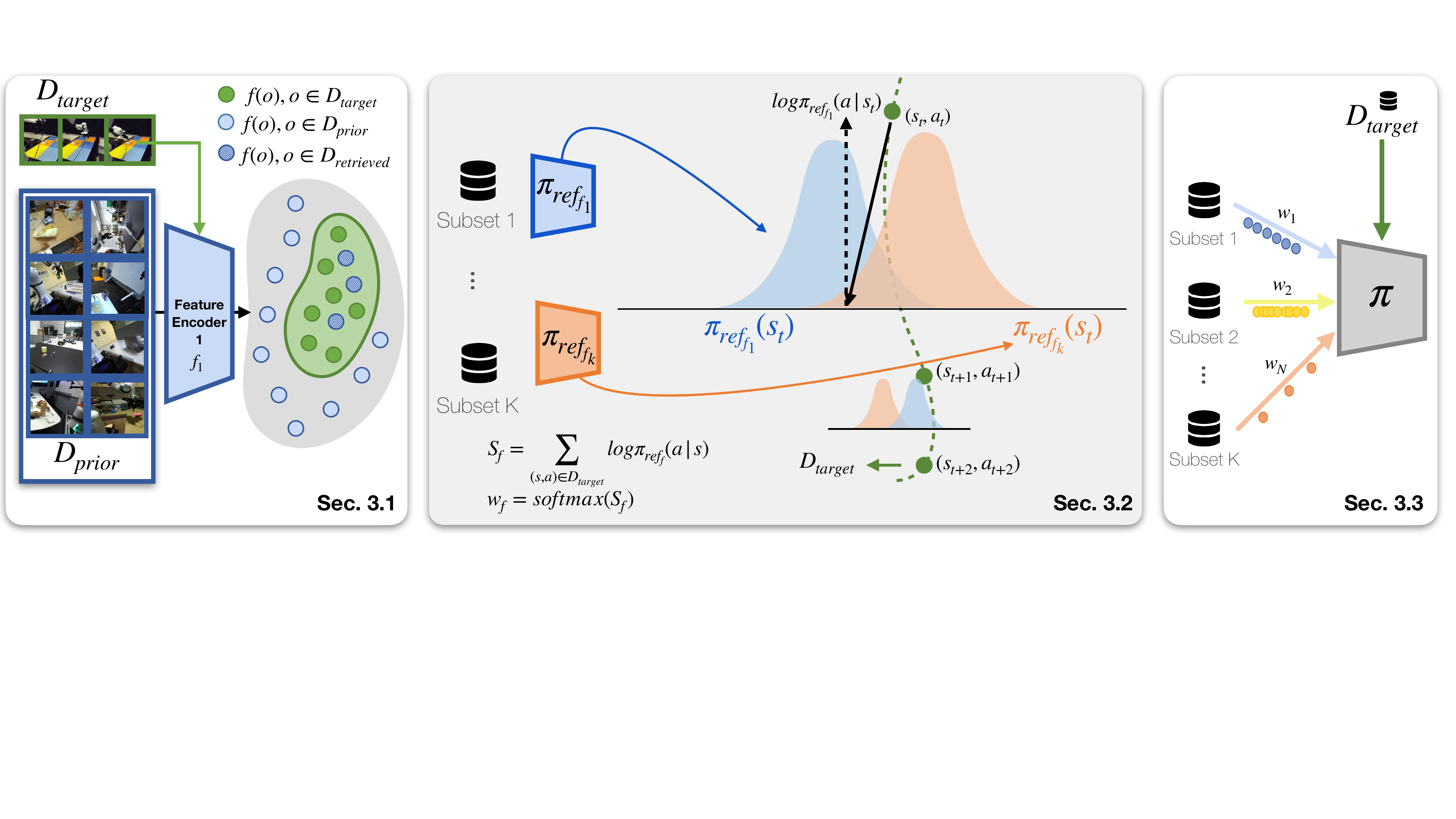}
    \vspace{-80pt}
    \caption{\textbf{Overview of our proposed COLLAGE approach.}
    Left: Given a set of target demonstrations $\dataset{target}$, each modality $f$ selects a set of retrieved trajectories $\dataset{retrieved}^{f}$ from a prior dataset $\dataset{prior}$.
    Center: We use the retrieved trajectories for each modality to train a reference policy $\pi_{\text{ref}_f}$.
    For each reference policy, we compute the log-likelihood $\log \pi_{\text{ref}_f}(a_t|s_t)$ of each state $s_t$ of the target dataset.
    With this, COLLAGE estimates adaptively the importance $w_{f}$ of each modality.
    Right: We train our final policy using the retrieved examples of each modality.
    We perform importance sampling, such that the sampling probability for each retrieved set is equal to $w_f$.
     }
    \label{fig:method}
\vspace{-1em}
\end{figure}

\textbf{Overview.}
Our proposed approach, \methodname{},
formulates this challenge as a retrieval problem: we aim to identify demonstrations from $\dataset{prior}$ that are most relevant to those in $\dataset{target}$ and are therefore more likely to support effective policy learning for the target task.
Rather than relying on a single-modality similarity measure, as in prior work, \methodname{} leverages multiple modalities to retrieve complementary and diverse demonstrations.

More specifically, in our setting, each state $s$ consists of multi-modal observations $o = \{I, d, r\}$, where $I$ is the RGB image, $d$ is the depth map, and $r$ is the robot state (e.g., joint angles or end-effector pose). We assume access to $k$ feature encoders $\{\mathcal{F}_1, \mathcal{F}_2, \cdots, \mathcal{F}_k\}$, where each feature encoder $\mathcal{F}_i$ corresponds to a different modality (e.g., vision, motion, shape), and maps an observation $o$ to a low-dimensional embedding suitable for computing similarity.

Our approach first retrieves subsets of relevant trajectories $\{ \dataset{retrieved}^{f_1}, \cdots, \dataset{retrieved}^{f_k} \}$ by using each feature encoder to measure similarity between demonstrations in the target and prior datasets (Figure~\ref{fig:method}, left). Then, we estimate an importance weight $w_f$ for each retrieved subset (Figure~\ref{fig:method}, center). Finally, we train a language-conditioned policy using the union of the target dataset and all retrieved datasets.
During policy training, we sample trajectories following the estimated importance weights $w_f$. 
(Figure~\ref{fig:method}, right).

\subsection{Retrieval Across Multiple Modalities}
In this section, we describe how to obtain the retrieved set for each feature modality.

\textbf{Granularity of Data Retrieval.} 
Before retrieving relevant data, we must decide the level of granularity for retrieval: individual states, sub-trajectories, or entire trajectories from $\dataset{prior}$. 
Taking inspiration from~\cite{memmel2024strap}, we primarily adopt sub-trajectory retrieval, which enables more flexible and semantically meaningful matching. 
However, our approach is agnostic to this choice, and in our instantiation of \methodname{}, we also incorporate a trajectory-level retrieval feature (see Section~\ref{sec:features}).

\textbf{Sub-trajectory based Data Retrieval.} Following~\citet{memmel2024strap}, we first segment each demonstration in $\dataset{target}$ into sub-trajectories using an action-based heuristic based on end-effector velocity.
For each feature modality, we use the corresponding encoder $\mathcal{F}_i$ and perform retrieval independently.
Given a segmented target sub-trajectory $t'$, and a trajectory $t$ from $\dataset{prior}$, we compute a pairwise cost matrix $C \in \mathbb{R}^{|t'| \times |t|}$ where $C_{ij} = \| \mathcal{F}_i(O_i) - \mathcal{F}_i(O_j) \|_2$.
We then apply Subsequence Dynamic Time Warping (S-DTW) to align $t'$ with a contiguous sub-sequence of $t$.
For each target sub-trajectory, we retrieve the top-$K$ lowest-cost matches from $\dataset{prior}$ according to S-DTW. Readers are referred to Section \ref{sec:S_DTW_formal} and~\cite{giorgino2009computing} for formal definitions of DTW and S-DTW.

This retrieval process is repeated separately for each feature modality, resulting in a set of modality-specific retrieval datasets $\{\dataset{retrieved}^{f_1}, \cdots, \dataset{retrieved}^{f_k}\}$.
Each retrieved sub-trajectory also inherits its corresponding language instruction.

\subsection{Estimating the Weights for Retrieved datasets}
\label{sec:weight}
Given the modality-specific retrieved datasets ${\dataset{retrieved}^{f_1}, \cdots, \dataset{retrieved}^{f_k}}$, a simple strategy, adopted in prior work~\citep{du2023behavior, lin2024flowretrieval, memmel2024strap}, is to uniformly combine them and perform augmented policy learning. However, as we demonstrate in our experiments, this naive uniform merging often leads to suboptimal performance. This is because, for a given task, some modalities retrieve more informative examples than others (see Figure~\ref{fig:teaser}).

To address this, we propose an adaptive weighting strategy that draws inspiration from importance sampling \cite{tokdar2010importance}. In this view, each retrieval modality $f$ induces a proposal distribution over examples in $\dataset{prior}$, and our goal is to sample examples from $\dataset{retrieved}^f$ that reflect how well these proposals approximate the target task distribution. 
Concretely, for each modality $f$, we train a lightweight behavior cloning (BC) policy $\pi_{\text{ref}_f}$ using only the modality-specific retrieved data $\dataset{retrieved}^{f}$:
\begin{align} \pi_{\text{ref}_f} = \arg\min_{\theta} \mathcal{L}_{\text{BC}}\left(\theta;\dataset{retrieved}^{f}\right). \end{align}
Then, we evaluate the relevance of each modality by computing the log-likelihood of the target demonstrations under the corresponding reference policy:
\begin{align} S_f = \sum_{(s, a, \ell) \in \dataset{target}} \log \pi_{\text{ref}_f}(a \mid s, \ell) \end{align}
These scores are normalized using a softmax function to produce a set of modality weights: $w_f = \frac{\exp(S_f / \tau)}{\sum_{f'} \exp(S_{f'} / \tau)}$.

Intuitively, a modality that retrieves more relevant data, --i.e., one whose reference policy better explains the target task-- receives a higher weight, analogous to a higher importance weight in importance sampling. The softmax normalization ensures that the weights are positive and sum to one, enabling us to automatically prioritize the most useful retrievals during training. 

\subsection{Retrieval Augmented Policy Learning} 
The next step is to
train a language-conditioned visuomotor policy \( \pi \) using behavior cloning. The policy is a transformer-based model~\cite{vaswani2017attention}, and takes as input the past \( h \) observations \( s_{t-h:t} \) along with a task instruction \( \ell \), predicting \( h \) future actions using a Gaussian Mixture Model (GMM) head. Following recent approaches~\cite{haldar2024baku, nasiriany2024robocasa}, our training objective combines multi-step prediction loss with $\ell_2$ regularization on model parameters $\theta$.
\begin{equation}
\label{eq:loss}
\mathcal{L}(\theta) = \mathbb{E}_{(s_{t-h:t},\ a_{t:t+h},\ \ell) \sim \mathcal{D}} \left[ 
    -\log\, \pi_\theta(a_{t:t+h} \mid s_{t-h:t},\ \ell) 
\right] + \lambda \|\theta\|_2^2
\end{equation}
\textbf{Sampling Strategy.}  
For each modality \( f \), we construct an augmented dataset \( \mathcal{D}^{f}_{\text{aug}} = \dataset{{retrieved}}^{f} \cup \dataset{{target}} \). During training, we sample examples from the augmented datasets $\mathcal{D}^{f}_{\text{aug}}$, 
using the sampling weights $w_{f}$ we estimated in Section~\ref{sec:weight}. This results in training batches containing examples retrieved by different modalities, biased toward those deemed more useful for the target task.
 \subsection{Feature Modalities for Data Retrieval}
\label{sec:features}
We instantiate \methodname{} using four feature modalities to measure similarity between $D_{\text{target}}$ and $D_{\text{prior}}$: \emph{visual}, \emph{motion}, \emph{shape}, and \emph{language}, derived from DINOv2~\cite{oquab2023dinov2}, Optical Flow~\cite{beauchemin1995computation}, PointNet~\cite{qi2017pointnet++}, and OpenAI embeddings~\cite{openai_embeddings_2023}, respectively. Below, we outline how features are extracted for a single state $s_i$ from a demonstration $d$; additional implementation details are in Sections~\ref{sec:sim_retrieval_implementation_details} and~\ref{sec:real_retrieval_implementation_detail}.

\textbf{Visual Similarity Feature.}  
We use DINOv2~\cite{oquab2023dinov2} to extract visual features from the RGB image $I_i$ of each state. The visual feature is defined as $\mathcal{F}_v(s_i) = \text{DINO}(I_i)$.

\textbf{Motion Similarity Feature.}  
We use Optical Flow~\citep{beauchemin1995computation} to measure motion similarity. Unlike DINO features, which can be extracted using a pretrained model, this modality requires a custom pretraining stage to reduce the dimensionality of raw optical flow. To this end, we first extract optical flow for the entire $\dataset{prior}$ dataset using GMFlow~\cite{xu2022gmflow}. The optical flow $o_i$ is computed between each image $I_i$ of $s_i$ and $I_{i+j}$ of $s_{i+j}$, where $j$ is determined based on the temporal offset used in the downstream behavior cloning (BC) policy. This results in the dataset $\text{FLOW}_{\text{prior}} = \{\mathcal{O}(I_i, I_{i+j}) \mid (s_i, a_i) \in \dataset{prior} \}$, where $\mathcal{O}$ denotes the optical flow operator, i.e., GMFlow in our case. Following ~\citet{lin2024flowretrieval}, we train an encoder $p_{\theta}(o_i)$ and a decoder $q_{\phi}(\cdot)$ using a reconstruction loss defined as $L(\theta, \phi) = \mathbb{E}_{o_i \sim \text{FLOW}_{\text{prior}}} \left\| q_{\phi}(p_{\theta}(o_i)) - o_i \right\|$. The encoder-decoder model is trained on $\dataset{prior}$, and as shown by~\citet{lin2024flowretrieval}, given a sufficiently large $\dataset{prior}$, the learned representation generalizes well to $\dataset{target}$. Finally, the motion feature is defined as $\mathcal{F}_m(s_i) = p_{\theta}(o_i)$.

\textbf{Shape Similarity Feature.}  
To represent the 3D geometry of the scene, we convert the depth map $m_i$ into a point cloud $\mathbf{p}_i$ after background segmentation. The point cloud is then passed through a pretrained PointNet++~\cite{qi2017pointnet++} model to compute the shape feature: $\mathcal{F}_p(s_i) = \text{PointNet}(\mathbf{p}_i)$.

\textbf{Language Similarity Feature.}  
We use OpenAI embeddings~\cite{openai_embeddings_2023} to encode the instruction $l_d$ associated with each demonstration. The language feature is computed as $\mathcal{F}_l(l_d) = \text{OpenAIEmbed}(l_d)$.
\section{Experiments}
\begin{figure}[!t ]
\centering
\scriptsize
\includegraphics[width=1.0\linewidth]{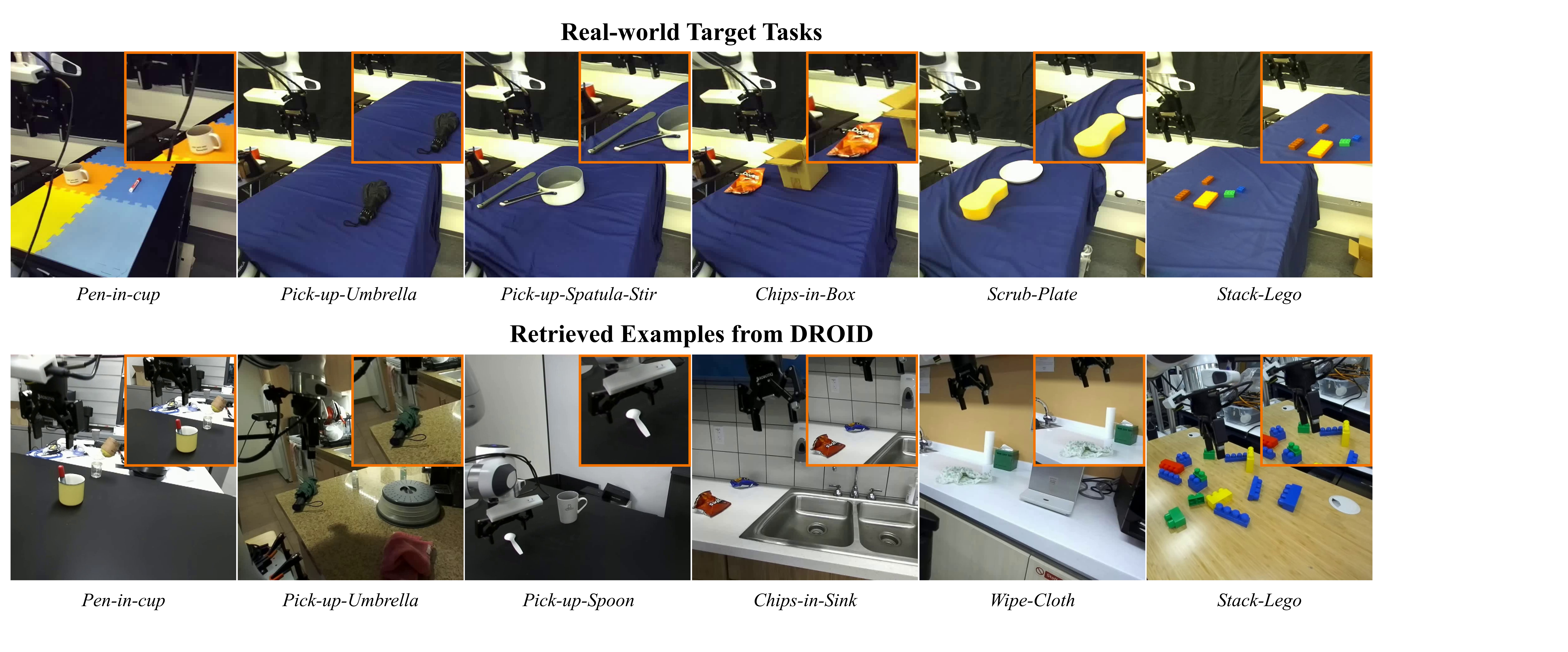}
\caption{\textbf{Real-world tasks and corresponding retrieved examples from DROID.}
Our real-world setting (top) is visually very different from DROID (bottom), yet we are able to retrieve meaningful and helpful demonstrations for each task.
In many cases our target task does not align perfectly with DROID examples, but we are still able to retrieve a relevant demonstration (e.g., wipe-cloth demonstration for the related scrub-plate task), or return a partial motion of a more complicated task (e.g., pick-up-spoon is relevant for the pick-up-spatula segment of the pick-up-spatula-stir task).
}
\label{fig:tasks}
\end{figure}

We evaluate our method on the simulated LIBERO~\cite{liu2023libero} benchmark. Additionally, we also test our method in the real world on \numtasks manipulation tasks using the DROID dataset~\cite{khazatsky2024droid}. Below, we provide a brief description of the tasks evaluated in this work which are also shown in Figure \ref{fig:tasks}.

\subsection{Experimental Setup}
\textbf{Tasks \& Datasets.} For our simulated experiments, we use $5$ demonstrations from each of the 10 tasks in the \emph{LIBERO-$10$} benchmark as $D_{target}$ and $4500$ demonstrations from \emph{LIBERO-$90$} as $D_{prior}$. 

For real-world evaluations, we use a FRANKA Panda robot.
Our $\dataset{prior}$ is a set of $30$k successful episodes from the DROID dataset~\cite{khazatsky2024droid} with language instruction annotations.
These episodes are selected randomly from a total of $50$k successful episodes in DROID.
This setting is particularly challenging, as the DROID demonstrations differ significantly in visual appearance from our robot setup (see Figure~\ref{fig:tasks}). Unlike previous works~\cite{du2023behavior, lin2024flowretrieval, memmel2024strap}, we do not collect the demonstrations for $\dataset{prior}$ ourselves and rely on automatic retrieval. We evaluate on \numtasks real-world tasks and collect \emph{5} demonstrations per task. The tasks are: (1) \emph{Pen-in-Cup}, the robot needs to pick up a pen and put it inside the cup; (2) \emph{Umbrella}, the robot needs to pick up the umbrella; (3) \emph{Spatula-Stir}, the robot needs to stir a pot using a spatula; (4) \emph{Chips-Box}, the robot needs to pick up the box of chips and put it inside the box, (5) \emph{Scrub-Plate}, where the robot scrubs a plate using a sponge and (6) \emph{Stack-Lego}, where the robot stacks one Lego 
block on top of another. All these tasks require fine-grained manipulation and are hard to learn directly from $5$ demonstrations. Notably, while some tasks are directly represented in DROID (e.g., \emph{Pen-in-Cup}), others, such as \emph{Scrub-Plate}, are not explicitly demonstrated, to the best of our knowledge. The tasks are selected to span a diverse set of behaviors, highlighting the importance of retrieving data using multiple modalities. 

\noindent\textbf{Baselines and Ablations.} 
We compare \methodname{} against: \textbf{Non-retrieval methods:} 
(1) \textbf{BC}: training a transformer-based policy using only $\dataset{{target}}$; 
(2) \textbf{MT}: training a multi-task transformer policy on $\dataset{{target}} \cup \dataset{{prior}}$; \textbf{Retrieval-based methods:} 
(3) \textbf{BR}~\cite{du2023behavior}: cosine similarity retrieval in a VAE-based state-action embedding space trained on $\dataset{{prior}}$; 
(4) \textbf{FR}~\cite{lin2024flowretrieval}: same as BR, but with a VAE trained on optical flow
(5) \textbf{STRAP}~\cite{memmel2024strap}: sub-trajectory retrieval using S-DTW with DINO features (numbers reported from~\cite{memmel2024strap}); 
(6) \textbf{Motion}: sub-trajectory retrieval using optical flow for motion similarity; 
(7) \textbf{Language}: retrieval at the demonstration level using OpenAI language embeddings.
(8) \textbf{Shape}: sub-trajectory retrieval using PointNet for 3D shape similarity; 
(9) \textbf{Visual}: STRAP reproduced by us.
\textbf{Ablation:} 
(10) \textbf{Non-Adaptive (NA)-Fusion}: uniform sampling across all retrieved demonstrations,
instead of using per-modality adapted weights. 
More details in Appendix \ref{sec:sim_details}  and \ref{sec:real_details}. 

\subsection{Results}
\label{sec:evals}
\begin{table}[!t]
    \scriptsize
    \centering
     \renewcommand{\arraystretch}{1.3}
\begin{tabular}{lcccccccccc}
    \toprule
    \textbf{Tasks} & \textbf{Mug-M} & \textbf{Book} & \textbf{Cheese} & \textbf{Soup-Sa} & \textbf{Mug-Mi} & \textbf{Soup-C} & \textbf{Bowl} & \textbf{Stove} & \textbf{Mug-P} & \textbf{Avg.} \\
    \midrule
    BC & $36.0$ & $44.7$ & $47.3$ & $17.3$ & $25.3$ & $19.3$ & $72.0$ & \underline{$73.3$} & {$12.7$} & $34.8$ \\
    MT & $31.3$ & $\mathbf{89.3}$ & $33.3$ & $15.3$ & $21.0$ & $23.3$ & $66.7$ & $68.0$ & $9.3$ & $35.8$ \\  
    \midrule
    BR~\cite{du2023behavior} & $32.7$ & $38.0$ & $33.3$ & $29.3$ & $16.3$ & $28.0$ & $74.7$ & $63.3$ & $6.7$ & $32.2$ \\
    FR~\cite{lin2024flowretrieval} & $28.7$ & $58.0$ & $29.3$ & $17.3$ & $21.3$ & $23.3$ & $83.3$ & $71.3$ & $16.7$ & $34.9$ \\
    STRAP~\cite{memmel2024strap} & $\mathbf{57.3}$ & $85.3$ & $29.3$ & $16.7$ & $29.3$ & $\mathbf{42.7}$ & $91.3$ & $\mathbf{85.3}$ & $\underline{18.7}$ & $45.6$ \\
    \midrule
    Motion & $48.0$ & $66.7$ & $60.7$ & $25.3$ & $28.0$ & $30.0$ & $48.0$ & \underline{$73.3$} & $15.3$ & $39.5$ \\
    Language & $34.7$ & $61.3$ & $66.0$ & $32.0$ & $20.7$ & $31.3$ & $93.3$ & $58.0$ & $13.3$ & $41.1$ \\
    Shape & \underline{$54.0$} & $77.3$ & $\mathbf{68.7}$ & $28.7$ & $20.7$ & $28.7$ & $94.0$ & $63.3$ & $12.7$ & $44.8$ \\
    
    Visual ~\cite{memmel2024strap} & $40.7$ & $66.0$ & $66.7$ & $30.0$ & \underline{$31.3$} & $29.3$ & $95.3$ & \underline{$73.3$} & $\mathbf{21.3}$ & $45.4$ \\    
    
    NA-Fusion & \underline{$54.0$} & \underline{$86.7$} & $55.3$ & \underline{$42.7$} & $24.7$ & $28.7$ & $94.7$ & $67.3$ & $14.7$ & $46.9$ \\
    
    \rowcolor{gray!20} \methodname{} & $51.3$ & $\mathbf{89.3}$ & \underline{$67.3$} & $\mathbf{53.3}$ & $\mathbf{32.7}$ & \underline{$33.3$} & $\mathbf{96.0}$ & $68.7$ & $13.3$ & $\mathbf{50.5}$ \\
    \bottomrule
\end{tabular}
    \caption{\textbf{Performance comparison on the \emph{LIBERO-10} benchmark.} 
    We compare \methodname{} with various baselines (first two groups) and with ablations of our approach (third group).
    \textbf{Bold} indicates the best performance, and \textbf{\underline{underline}} indicates the second-best performance. All sub-trajectory method results are reported with $K=100$ sub-trajectories. Results are averaged over $3$ seeds and $50$ trials. The last column reports the average across all $10$ tasks in \emph{LIBERO-10}, including the Moka-Moka task, for which all methods achieve a success rate of $0$. Individual success rates are reported only for the remaining $9$ tasks. }
    \label{tab:sim_performance}
\end{table}

Our experimental evaluations aim to answer the following 
questions:

(1) \emph{Can we improve imitation learning performance by combining multiple feature modalities?}
Our evaluation on the \emph{LIBERO-10} benchmark (Table \ref{tab:sim_performance}) shows that \methodname{} improves average performance by $5.1\%$  and $5.7\%$ over \emph{Visual} and \emph{Shape} based retrieval respectively. 
While \emph{Visual} is the strongest single-modality baseline overall, it often fails to retrieve demonstrations 
in cases with significant appearance changes.
 For example, on the \emph{Book} task, \emph{Visual} modality based retrieval does not return demonstrations in $\dataset{{prior}}$ that share the same instruction but differ in the visual appearance of the scene. 
In contrast, \methodname{} retrieves such demonstrations since it also relies on the \emph{Motion} and \emph{Language} modalities, resulting in a $23.3\%$ improvement over \emph{Visual} modality alone. 
Similarly, \emph{Motion}-based retrieval underperforms on many tasks because it ignores semantic information which is critical for some tasks. 
Across the benchmark, \methodname{} also consistently outperforms prior retrieval baselines, such as \emph{BR} and \emph{FR}, in imitation learning performance.

(2) \emph{Can \methodname{} effectively retrieve relevant demonstrations from large-scale, diverse datasets that have minimal overlap to the target tasks?} Prior works~\cite{du2023behavior, lin2024flowretrieval, memmel2024strap} demonstrate data retrieval by collecting a set of $\dataset{{prior}}$ demonstrations that are relatively similar to the target demonstrations. In contrast to that, we go one step further and tackle the challenging setting of using as $\dataset{{prior}}$ an independent large-scale dataset, \emph{DROID}. From our experiments (Table~\ref{tab:real_performance}), we observe that
\emph{BC} performs poorly in the real-world setting, with just $0.5$/$15$ successes on average--highlighting the difficulty of learning fine-grained behaviors from limited demonstrations. 
Individual modalities perform well on specific tasks (e.g., \emph{Language} on \emph{Scrub-Plate}, \emph{Shape} on \emph{Stack-Lego}, \emph{Visual} on \emph{Chips-Box}), but \methodname{} consistently outperforms all of them. We find that adaptively sampling the per-modality demonstrations improves robustness to real-world variations such as lighting and object pose. Interestingly, \methodname{} can outperform the best individual modalities by combining complementary behaviors across modalities (e.g. \emph{Spatula-Stir}, \methodname{} $9/15$ vs \emph{Language} $5/15$) . 
 
(3) \emph{How important is adaptive weighting when using data retrieved from multiple modalities?} 
As shown in Table~\ref{tab:sim_performance}, \methodname{} outperforms the non-adaptive baseline (NA-Fusion) across all tasks (50.5 vs.\ 46.9 average success rate). We find that the importance weights predicted by \methodname{} (Table \ref{tab:modality_weights}) are high for modalities with high success rates (e.g., \emph{Shape} on \emph{Cheese-Butter} and \emph{Motion} on \emph{Soup-Cheese}). Similarly, for real-world tasks, \methodname{} down-weights the less informative modality (e.g., \emph{Motion} for Lego and \emph{Shape} for Pen-in-Cup).
 
\begin{table}[]
\centering
\begin{minipage}[t]{0.48\textwidth}
\scriptsize
\renewcommand{\arraystretch}{1.1}
\setlength{\tabcolsep}{2.0pt}
\begin{tabular}{lccccccc}
\toprule
\textbf{Method} & \textbf{Pen} & \textbf{Umb} & \textbf{Stir} & \textbf{Chips} & \textbf{Lego} & \textbf{Scrub} & \textbf{Avg.}\\
\midrule
    BC & $0 / 15$ & $2 / 15$ & $1/15$ & $3/15$ & $0/15$ & $0/15$ &$6.7$\\
    MT &  $0/15$ & $0/15$ & $0/15$ & $1/15$ & $2/15$ & $0/15$ &$3.3$\\
    \midrule
    Visual \cite{memmel2024strap} & $4/15$ & $4/15$ & $3/15$ & $\mathbf{9/15}$& $5/15$ & $1/15$ & $28.9$\\
    Motion & $0/15$ & $0/15$ & $4/15$ & $\mathbf{9/15}$ & $1/15$ & $1/15$ &$16.7$\\
    Shape & $0/15$ & $3/15$ & $3/15$ & $6/15$ & $\mathbf{6/15}$ & $0/15$ & $20.0$\\
    Language & $4/15$ & $4/15$ & $5/15$ & $4/15$ & $2/15$& $6/15$ & $27.7$\\
    \rowcolor{gray!20} \methodname{} &  $\mathbf{6/15}$ & $\mathbf{6/15}$ & $\mathbf{9/15}$ & $7/15$ & $\mathbf{6/15}$ & $\mathbf{7/15}$ & $\mathbf{45.5}$\\
\bottomrule
\end{tabular}
\caption{\textbf{Real-world results.} We perform 15 trials for each method and we report the number of successful episodes for each task. The last column refers to the average percentage of success. \textbf{Bold} indicates the best performance.}
\label{tab:real_performance}
\end{minipage}
\hfill
\begin{minipage}[t]{0.42\textwidth}
\scriptsize
\centering
\renewcommand{\arraystretch}{1.05}
\begin{tabular}{lcccc}
\toprule
\textbf{Task} & \(w^{\text{Visual}}\) & \(w^{\text{Motion}}\) & \(w^{\text{Shape}}\) & \(w^{\text{Language}}\) \\
\midrule
\multicolumn{5}{c}{\textbf{LIBERO-10}} \\
\midrule
Cheese & $0.28$ & $0.18$ & $0.46$ & $0.07$ \\
Soup-C & $0.11$ & $0.52$ & $0.22$ & $0.14$ \\

\midrule
\multicolumn{5}{c}{\textbf{DROID}} \\
\midrule
Lego & $0.6$ & $0.02$ & $0.28$ & $0.1$\\ 
Pen & $0.30$ & $0.24$ & $0.10$ & $0.36$ \\
\bottomrule
\end{tabular}
\caption{\textbf{Importance weights for various tasks.} We report the weights for each modality for simulation (LIBERO-10) and real-world (DROID) tasks.}
\label{tab:modality_weights}
\end{minipage}
\vspace{-11pt}
\end{table}  

\section{Conclusion}
\label{sec:conclusion}
We propose \methodname{}, an approach that enables robust retrieval-augmented policy learning by adaptively combining data retrieved using different similarity measures. Unlike prior approaches that rely on fixed feature types or manually designed retrieval pipelines, \methodname{} is agnostic to the choice of retrieval modality, making it broadly applicable across diverse settings. Our results show that adaptively weighting data retrieved from diverse similarity modalities improves few-shot imitation learning performance, both in simulation and when retrieving from large-scale, visually diverse datasets like DROID. We will publicly release our code to support further research.

\section{Limitations}
While \methodname{} provides a flexible and general approach for combining data retrieved via multiple modalities, it has a few practical limitations. The performance of our method ultimately depends on the quality of the retrieved data—if all modalities fail to retrieve relevant examples, performance gains may be limited. However, we find this to be rare in practice, especially when using diverse modalities that offer complementary strengths. Additionally, our approach requires training a reference policy for each modality, which, although lightweight, introduces additional computational overhead. 

An exciting direction for future work is to explore more efficient ways to approximate these reference policies or develop proxy metrics that can predict the utility of retrieved data without full policy training.

\acknowledgments{ 
We thank Jiaheng Hu and Albert Yu for their valuable feedback on the manuscript, and appreciate fruitful discussions with Arpit Bahety, Rutav Shah, Marius Memmel, Jacob Berg, and Sanjay Haresh. G.P. acknowledges support by NSF IIS-2504906, and Gifts from Adobe and Google. R.M.M. acknowledges support by DARPA TIAMAT HR0011-24-9-0428.}

\bibliography{bibliography} 
\newpage
\appendix
\section{Appendix}
\label{s:appendix}

\subsection{Details of Simulated Experiments}
\label{sec:sim_details}
\textbf{Full Task Names for Table 1 in the main paper} Below are the target abbreviations used in Table 1 and Table \ref{tab:libero10_stds} respectively and the corresponding task name as used in \emph{LIBERO-10} benchmark:

\begin{table}[h!]
\centering
\tiny
\renewcommand{\arraystretch}{1.2}
\begin{tabular}{lll}
\toprule
\textbf{Abbr. (Tab. \ref{tab:sim_performance})} & \textbf{Abbr. (Tab. \ref{tab:libero10_stds}, Fig. \ref{fig:weights})} & \textbf{Task Name in LIBERO-10} \\
\midrule
Mug-M   & Mug-Mug& \tiny {LIVING ROOM SCENE5 put the white mug on the left plate and put the yellow and white mug on the right plate}\\
Book    & Book-Caddy  & \tiny{STUDY SCENE1 pick up the book and place it in the back compartment of the caddy} \\
Cheese  & Cheese-Butter & \tiny{LIVING ROOM SCENE2 put both the cream cheese box and the butter in the basket} \\
Soup-Sa & Soup-Sauce  & \tiny{LIVING ROOM SCENE2 put both the alphabet soup and the tomato sauce in the basket} \\
Soup-C  & Soup-Cheese & \tiny{LIVING ROOM SCENE1 put both the alphabet soup and the cream cheese box in the basket} \\
Stove & Stove-Moka & \tiny{KITCHEN SCENE3 turn on the stove and put the moka pot on it} \\
Mug-Mi  & Mug-Micro   & \tiny{KITCHEN SCENE6 put the yellow and white mug in the microwave and close it}\\
Mug-P   & Mug-Pudding & \tiny{LIVING ROOM SCENE6 put the white mug on the plate and put the chocolate pudding to the right of the plate} \\
Bowl    & Bowl-Cabinet & \tiny{KITCHEN SCENE4 put the black bowl in the bottom drawer of the cabinet and close it} \\
\bottomrule
\end{tabular}
\caption{Mapping between abbreviations in Tables \ref{tab:sim_performance} and \ref{tab:libero10_stds} and the full task names.}
\label{tab:task_abbr_mapping}
\end{table}

\textbf{Retrieval Implementation Details}
\label{sec:sim_retrieval_implementation_details}
We perform retrieval using $4$ feature modalities i.e. Visual using DINO features \cite{oquab2023dinov2}, Shape using POINTNET features \cite{qi2017pointnet++}, Motion using Optical Flow features \cite{xu2022gmflow}  and Language using embeddings obtained using OPENAI's API \cite{openai_embeddings_2023}. We set the number of retrieved sub-trajectories, \emph{K} to 100, for DINO, PointNet and FLOW respectively. We use the agent-centric images for all modalities.
Below, we describe the implementation pipeline of each feature in detail: 
\begin{itemize}
    \item \textbf{DINO}: We use the $128 \times 128$ images provided in the \emph{LIBERO-10} benchmark and extract embeddings using \emph{DINOv2}, specifically we use the DINOv2 model from the \emph{transformers} codebase developed by \emph{HuggingFace} \cite{wolf-etal-2020-transformers}. The model outputs patch-wise low-dimensional embeddings along with the \emph{CLS} token, we apply average pooling over both to obtain a $1 \times 768$ vector for each image frame.

    \item \textbf{FLOW}: We extract optical flow using the GMFlow model~\cite{xu2022gmflow}, with a temporal offset of $j = 5$. To train an optical flow reconstruction model, we design a hybrid architecture that combines convolutional layers with a transformer-based encoder-decoder. Given a $128 \times 128$ flow input, we first apply a series of convolutional operations to reduce the spatial dimensions and obtain a compact token of size $1 \times 512$. This token is passed through a transformer encoder, yielding a latent embedding of size $256$. The embedding is then processed by a transformer decoder and a sequence of convolutional upsampling layers to reconstruct the original optical flow, supervised using a reconstruction loss.

    \item  \textbf{PointNet}: We first extract the depth maps using the \emph{Robomimic}~\citep{mandlekar2021matters} codebase. Then, we use ground-truth segmentation masks obtained through the simulator to extract the foreground segmentation mask. This foreground segmentation mask is used to create the pointcloud of the scene. The pointcloud is passed through a PointNet++ model~\citep{qi2017pointnet++} to extract a $1 \times 256$ embedding vector for each observation. We use the PointNet++ model trained on the task of object classification using the \emph{ModelNet40}~\citep{wu20153d} dataset.

    \item \textbf{LANG}: We implement a language-based retrieval pipeline that leverages semantic similarity between task instructions. For each task, we compute a language embedding of the instruction using OpenAI’s \texttt{text-embedding-ada-002} model and compare it to embeddings from a large prior dataset using cosine similarity. Demonstrations with instruction similarity above 0.90 are selected as relevant. To ensure the number of retrieved data points matches that of the other modalities, we first compute the total number of data points retrieved by the visual modality and then distribute this budget evenly across the selected language-retrieved demonstrations. This enables a controlled comparison of retrieval performance across modalities under equal data budgets. Since we retrieve $100$ sub-trajectories for the other modalities, their dataset sizes remain very similar.

\end{itemize}

\textbf{Implementation Details of Segmenting $\dataset{target}$.} To perform sub-trajectory retrieval, we segment the demonstrations in $\dataset{target}$ using the heuristic proposed in ~\citep{memmel2024strap}. Specifically, we implement a trajectory segmentation heuristic based on the velocity magnitude of end-effector states. Given a sequence of 3D positions, we compute the frame-to-frame difference in the $(X, Y, Z)$ coordinates and sum the absolute values to approximate the velocity magnitude at each timestep. Points where this velocity falls below a small threshold $\epsilon$ (set to $5 \times 10^{-3}$ in simulated experiments) are interpreted as pauses in movement. The trajectory is then segmented at these pause points, producing a sequence of sub-trajectories that correspond to continuous motion segments between stops. This enables the isolation of meaningful motion primitives from longer trajectories. In practice, this strategy may lead to over-segmentation due to stop-motions in demonstrations, we fix this by merging all sub-trajectories with length greater than $20$.

\textbf{Baseline Implementation Details.} For BR ~\citep{du2023behavior} and FR ~\citep{lin2024flowretrieval}, we use the official codebase \footnote{\url{https://github.com/lihenglin/bridge\_training\_code}} from FlowRetrieval~\citep{lin2024flowretrieval} to train VAEs and compute similarities. We retrieve single state action pairs and pad them by retrieving states from $t-h$ to $h-1$ to ensure compatibility with the transformer based policy used in our experiments. For STRAP, we use the officially released codebases for retrieval \footnote{\url{https://github.com/WEIRDLabUW/STRAP}} and policy learning \footnote{\url{https://github.com/WEIRDLabUW/robomimic_strap}} respectively. 

\textbf{Policy Training Implementation Details.} 
\label{sec:policy_training_implementation_details}
All policies are trained with both agent and in-hand observation.  We use the Robomimic codebase\footnote{\url{https://github.com/ARISE-Initiative/robomimic/tree/robocasa}} to train our policies. Note that we keep our policy training and retrieval framework consistent with~\citet{memmel2024strap}. However, we find that their reported numbers (indicated as \emph{STRAP} in Table~\ref{tab:sim_performance} and Table~\ref{tab:libero10_stds}) do not exactly match the results we reproduced (indicated as \emph{DINO}). That said, the average success rate across all tasks is consistent with their reported value. After discussion with the authors of ~\citet{memmel2024strap}, we attribute this discrepancy primarily to the inherent stochasticity in the \emph{Robomimic} codebase.

\textbf{Hyperparameters.} For our implementation as well as baselines, we train a transformer based policy is trained for a total of $60k$ steps with a batch size of $32$. The number of target demonstrations is $5$ in all our experiments. The temperature parameter $\mathcal{T}$  introduced in Section \ref{sec:approach} is set to $2$ for all experiments in simulation. All reported success rates are averaged over 3 seeds ($1234$, $42$, $4325$).

\textbf{Results with Standard Deviations.} Table \ref{tab:libero10_stds} presents our complete results with standard deviations.

\begin{table}[h]
\scriptsize
\centering
\renewcommand{\arraystretch}{1.3}
\begin{tabular}{lccccc}
\toprule


& \textbf{Mug-Mug} & \textbf{Book-Caddy} & \textbf{Cheese-Butter} & \textbf{Soup-Sauce} & \textbf{Mug-Micro}   \\
\midrule
{BC} & $36.00 \pm 4.32$ & $44.67 \pm 3.40$ & $47.33 \pm 9.29$ & $17.33 \pm 3.40$ & $25.33 \pm 8.06$ \\
{MT} & $31.33 \pm 3.06$ & $89.33 \pm 5.89$ & $33.33 \pm 5.54$ & $15.33 \pm 3.46$ & $23.33 \pm 6.48$\\
\midrule
{BR} \cite{du2023behavior}&  $32.67\pm1.15$ & $38.00\pm2.0$ & $33.33\pm1.15$ & $29.33\pm3.06$ & $16.33\pm3.27$ \\


FR\cite{lin2024flowretrieval} & $28.67 \pm 4.19$ & $58.00 \pm 3.27$ & $29.33 \pm 4.62$ & $17.33 \pm 3.40$ & $21.33 \pm 3.06$ \\

STRAP \cite{memmel2024strap} &$57.33\pm7.7$ & $85.33\pm2.8$ & $29.33\pm11.3$ & $16.67\pm2.0$ &$29.33\pm2.7$ \\
\midrule
Motion & $48.00\pm4.24$ & $66.67\pm20.8$ & $60.67\pm2.86$ & $25.33\pm4.11$ & $28.00\pm9.93$\\
Language & $34.67\pm3.40$& $61.33\pm6.60$ & $66.00\pm9.09$ & $32.00\pm8.48$ & $20.67\pm1.89$\\
Shape & $54.00 \pm 5.89$ & $77.33\pm2.49$ & $68.67\pm15.43$ & $28.67\pm5.74$ & $20.67\pm4.71$\\
Visual \cite{memmel2024strap} & $40.67\pm3.06$ & $66.00 \pm3.46$ & $66.67\pm8.08$ & $30.00\pm3.46$ & $31.33\pm7.57$\\
NA-Fusion & $54.00\pm5.66$ & $86.67\pm4.99$ & $55.33\pm5.25$ & $42.67\pm5.25$ & $24.67\pm2.49$ \\
\methodname{} & $51.33\pm9.8$ & $89.33\pm4.99$&$67.33\pm10.49$ & {$53.33\pm8.38$}  & $32.67\pm5.30$\\
\bottomrule

\end{tabular}

\vspace{0.5em}

\begin{tabular}{lccccc}
\toprule
& \textbf{Soup-Cheese} & \textbf{Moka-Moka} & \textbf{Bowl-Cabinet} & \textbf{Stove-Moka} & \textbf{Mug-Pudding} \\
\midrule
BC & $19.33\pm2.49$&$00.00\pm0.00$ & $72.00\pm2.82$ & $73.33\pm4.98$ & $12.67\pm5.73$
 \\ 
MT & $23.33 \pm 3.06$ & $00.00 \pm 0.00$ & $66.67 \pm 4.19$ & $68.00 \pm 3.27$ & $9.33 \pm 3.06$ \\
\midrule
BR \cite{du2023behavior} & $28.00 \pm 4.62$ & $00.00 \pm 0.00$ & $74.67 \pm 3.77$ & $63.33 \pm 3.06$ & $6.67 \pm 3.40$ \\
FR \cite{lin2024flowretrieval} & $23.33 \pm 4.19$ & $00.00 \pm 0.00$ & $83.33 \pm 3.06$ & $71.33 \pm 3.40$ & $16.67 \pm 3.06$ \\

STRAP \cite{memmel2024strap} & $42.67\pm7.2$  & $00.00 \pm 0.00$ & $91.33\pm2.2$ & $85.33\pm2.2$ & $18.67\pm1.4$\\
\midrule
Motion & $30.00 \pm 7.11$& $00.00\pm0.00$& $48.00\pm4.24$ &$73.33\pm3.00$ & $15.33\pm3.4$ \\
Language &$31.33\pm7.72$  & $00.00\pm0.00$ & $93.33\pm3.4$ &$58.00\pm13.95$ & $13.33\pm2.49$\\
Shape & $28.67\pm5.24$ & $00.00\pm0.00$ & $94.00\pm1.63$ & $63.33\pm19.48$ & $12.67\pm9.43$\\
Visual \cite{memmel2024strap} &$29.33\pm4.6$ & $00.00\pm0.00$ & $95.33\pm2.49$ & $73.33\pm3.26$ & $21.33\pm2.49$\\
NA-Fusion & $28.67\pm5.73$ & $00.00\pm0.00$& $94.67\pm4.11$ & $67.33\pm19.96$ & $14.67\pm4.11$\\
 \methodname{} & $33.33\pm3.39$ & $00.00\pm0.00$ & $96.00\pm2.62$ & $68.67\pm16.36$ & $13.33\pm5.73$ \\

\bottomrule
\end{tabular}

\caption{Performance across 10 \emph{LIBERO-10} tasks with standard deviation reported across 3 seeds.}
\label{tab:libero10_stds}
\end{table}

\textbf{Comparison with higher values of retrieved sub-trajectories ($K$)} For \methodname{}, we retrieve \(K=100\) sub-trajectories per modality across four modalities, resulting in \(4 \times 100 = 400\) retrieved examples—compared to the \(100\) sub-trajectories used by a single-modality baseline. To isolate the effect of multi-modal information, we also ran an experiment retrieving \(K=400\) samples using only the best-performing modality, \emph{DINO}, matching our total retrieval count. The total number of sequences used on average by each method in this experiment is approximately $25k$. As shown in Table \ref{tab:size_ablation}, even with near identical retrieval size, \methodname{} outperforms the \emph{DINO}-only setup, demonstrating that combining multiple modalities yields benefits beyond merely increasing the number of examples.

\begin{table}[!t]
    \scriptsize
    \centering
    \renewcommand{\arraystretch}{1.3}
\begin{tabular}{lcccccccccc}
    \toprule
    \textbf{Tasks} & \textbf{Mug-M} & \textbf{Book} & \textbf{Cheese} & \textbf{Soup-Sa} & \textbf{Mug-Mi} & \textbf{Soup-C} & \textbf{Bowl} & \textbf{Stove-Mo} & \textbf{Mug-P} & \textbf{Avg.} \\
    \midrule

    Visual ($K=400$) & {$34.7$} & {$64.7$} & $\mathbf{72.0}$ & {$50.7$} & $26$  & $28.7$ & $95.3$ & $\mathbf{74.7}$ & $\mathbf{16.7}$ & $46.9$ \\
    \methodname{} & $\mathbf{51.3}$ & $\mathbf{89.3}$ & {$67.3$} & $\mathbf{53.3}$ & $\mathbf{32.7}$ & $\mathbf{33.3}$ & $\mathbf{96.0}$ & $68.7$ & $13.3$ & $\mathbf{50.5}$ \\
    \bottomrule
\end{tabular}
    \caption{Comparison between \emph{Visual} similarity only and \methodname{} while using identical retrieval set size on the \emph{LIBERO-10} benchmark. Both methods retrieve around $25k$ sequences with number of retrieved sub-trajectories, $K=400$. Results averaged over $3$ seeds. Best result in \textbf{bold}.}
    \label{tab:size_ablation}
\end{table}

\begin{figure}[!t]

    \centering
\includegraphics[width=\textwidth]{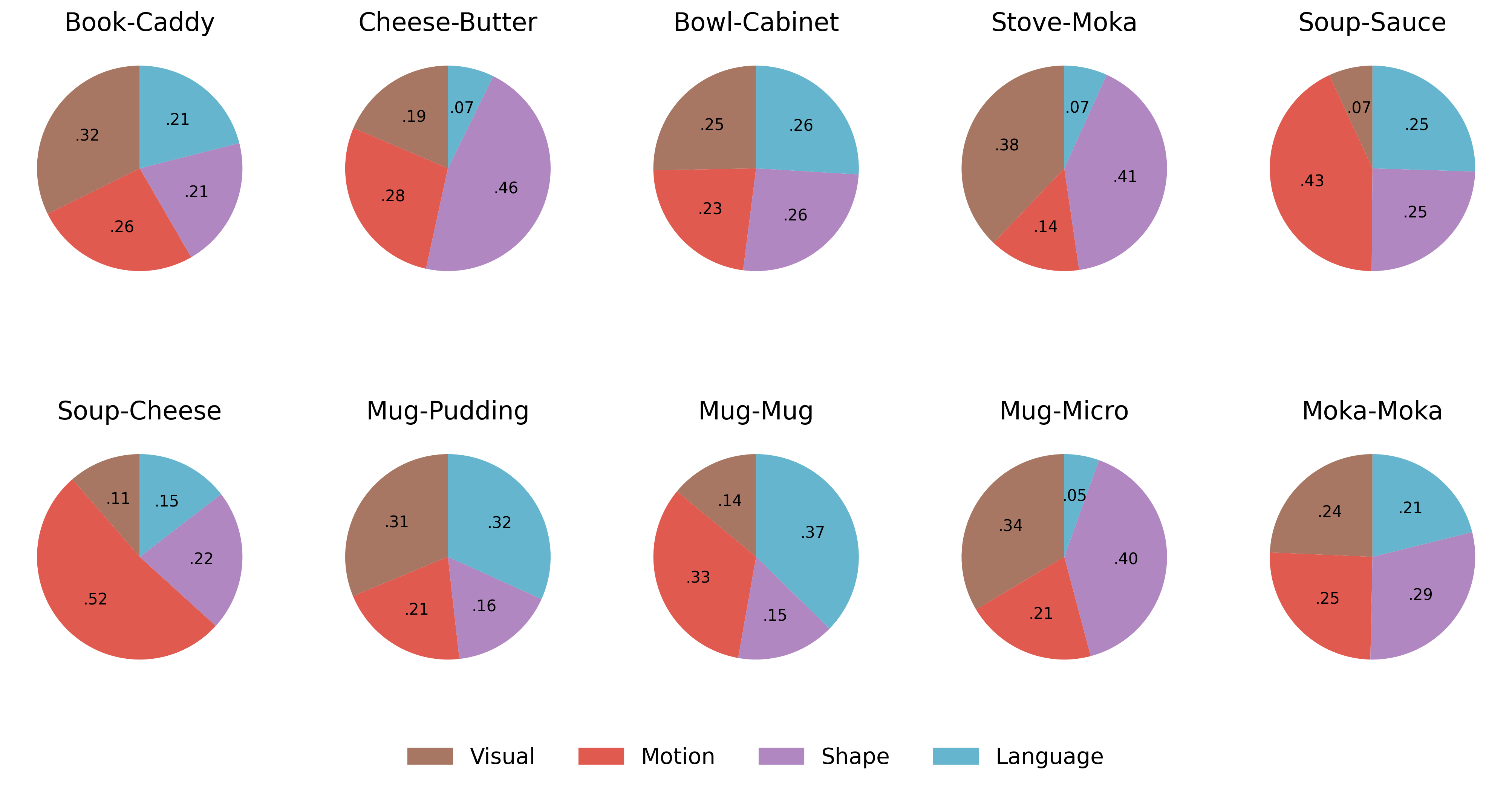}
\caption{Modality importance weights predicted by \methodname{} for all tasks in the \emph{LIBERO-10} benchmark.} 
\label{fig:weights}
\end{figure}

\textbf{Weights Predicted by \methodname{}.} Figure \ref{fig:weights} presents the weights predicted for each modality on all tasks from the \emph{LIBERO-10} benchmark.

\begin{figure}[!t]
    \centering
\includegraphics[width=\textwidth]{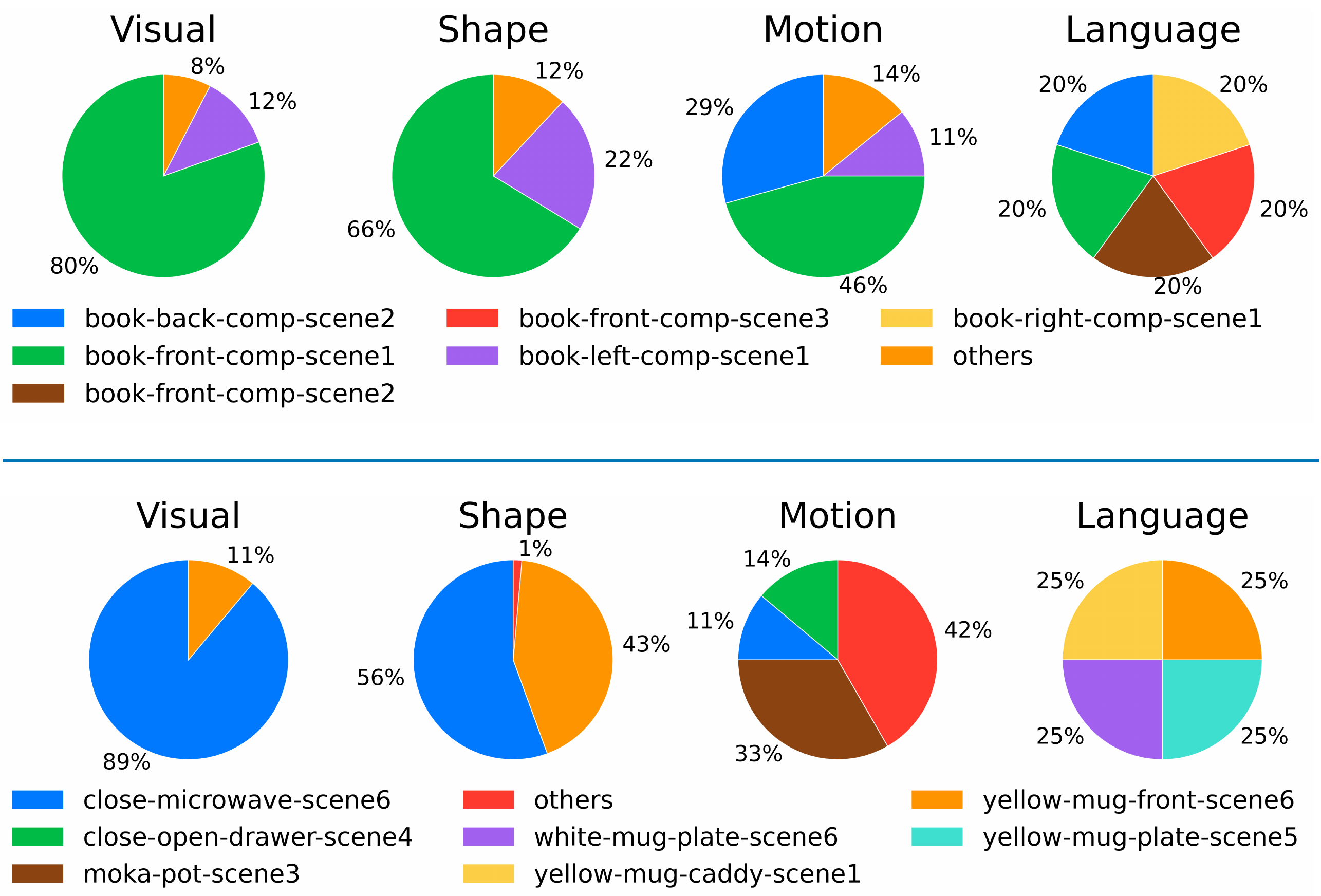}
\caption{Visualization of the types of demonstrations retrieved from \emph{LIBERO-90} for two \emph{LIBERO-10} target tasks: \emph{book-back-comp-scene1} (top) and \emph{mug-microwave-scene6} (bottom). Each pie chart corresponds to a retrieval modality and shows the proportion of retrieved sub-trajectories originating from different \emph{LIBERO-90} tasks. Tasks contributing less than $10\%$ of the retrieved data are grouped under “others”.}
\label{fig:distribution}
\end{figure}

\textbf{Distribution of Retrieved Data from Different Modalities.} In Figure~\ref{fig:distribution}, we visualize the demonstrations retrieved for the \emph{Book} (top) and \emph{Mug-Microwave} (bottom) tasks using the different modalities considered in this work. For the \emph{Book} task, visual retrieval primarily returns demonstrations from the same scene (\emph{scene1}) but with a different goal (e.g., placing the book in the front compartment). Because no retrieved example matches both the target scene and goal, motion and language retrieval contribute demonstrations with the correct goal but from a different scene (\emph{scene2}). Fusing these complementary modalities enables \methodname{} to outperform any individual modality, as shown in Table~1. For the \emph{Mug-Microwave} task, visual and shape retrieval yield demonstrations with relevant objects (e.g., yellow mug, microwave) and sub-tasks. In contrast, language retrieval surfaces demonstrations with semantically similar instructions but often from unrelated scenes or involving different objects. Accordingly, the modality weights predicted by \methodname{} assign a low weight (0.05) to \emph{Language} (see Figure \ref{fig:weights}), reflecting the limited task relevance of the retrieved examples.

\subsection{Details of Real World Experiments}
\label{sec:real_details}

\textbf{Retrieval Implementation Details}
\label{sec:real_retrieval_implementation_detail}
Our real-world implementation follows the same pipeline described in Section~\ref{sec:sim_retrieval_implementation_details} for retrieving data with the \emph{DINO}, \emph{FLOW}, and \emph{LANG} modalities. For the \emph{PointNet} modality, however, we no longer have simulator-provided ground-truth depth and masks. Instead, we start from the raw video files released with DROID~\cite{khazatsky2024droid}, which were captured using ZED stereo cameras. We use ZED’s Python API\footnote{\url{https://github.com/stereolabs/zed-python-api}} to extract per-frame depth maps. Next, we run DECOLA~\cite{cho2024language} to obtain object masks, which we combine with the depth maps to produce a foreground point cloud for each observation. Finally, this point cloud is fed into the PointNet++ model exactly as in Section~\ref{sec:sim_retrieval_implementation_details}.

\textbf{Camera Views.} We use all three available camera views in DROID for retrieval. For each modality, we obtain embeddings for each camera view and the embedding for a given observation is computed as the average of the embeddings from all three viewpoints.

\textbf{Hyperparameters.} The temperature parameter, $\mathcal{T}$ in real world experiments is set to $10$. The high value of temperature when compared to simulation experiments is mainly due to the large variance between DROID~\cite{khazatsky2024droid} data and the target demonstrations. The number of sub-trajectories retrieved, $K$, is set to $100$ in all real-world experiments. The threshold for segmenting trajectories, $\epsilon$, is set to $2 \times 10^{-3}$.

\textbf{Policy Training Implementation Details.} All baselines and our method is trained for $50k$ steps and the policies use all three camera views. The remaining policy implementation details match with those described for simulated experiments in Section \ref{sec:policy_training_implementation_details}. We use the setup from DROID for data collection and policy rollouts \cite{khazatsky2024droid}.

\subsection{Weight Estimation Implementation Details}

We train reference policies using a transformer-based architecture following the objective in Eq.~\ref{eq:loss}. Each policy is trained for 100 epochs (approximately one-third the rollout training time). To compute the log-likelihood score for modality \(f\), we first define the per-epoch evaluation:
\begin{align}
  S_f^{(e)}
  &= \sum_{(s,a,\ell)\in \mathcal{D}_{\mathrm{target}}}
     \log \pi_{\mathrm{ref}_f}^{(e)}\bigl(a\mid s, \ell\bigr).
\end{align}
We evaluate \(S_f^{(e)}\) every 10 epochs, excluding the first 50 epochs (i.e.\ at \(e\in\{60,70,80,90,100\}\)). The final modality score \(S_f\) is the average over these five checkpoints:
\begin{align}
\label{eq:relevance_weights}
   S_f
  &= \frac{1}{5}\sum_{e\in\{60,70,80,90,100\}} S_f^{(e)}.
\end{align}

We exclude the scores computed in the first 50 epochs because they exhibit high variance. We are interested in how well the retrieved data describes \(\mathcal{D}_{\mathrm{target}}\), which can only be evaluated after an initial training policy phase. Similarly, we do not train the policies beyond 100 epochs, as further training would cause overfitting to the retrieved data and yield non-meaningful scores.

\noindent\textbf{Computational Overhead of Weight Estimation.} Training each reference policy takes approximately 40 minutes, and all four are trained in parallel. Computing the relevance weights (Eq.~\ref{eq:relevance_weights}) requires only 1.5 additional minutes, owing to the small size of the target dataset. In comparison, training the final policy takes about 2 hours. This overhead is modest, especially considering that it results in performance improvements on 7 out of 9 tasks (\methodname{} vs.\ NA-Fusion, Table~1).

\subsection{Subsequence Dynamic Time Warping (S-DTW)}
\label{sec:S_DTW_formal}

\textbf{Dynamic Time Warping (DTW)} is a classical algorithm for computing an optimal alignment between two sequences that may differ in length or exhibit temporal misalignment. Given sequences \( X = \{x_1, \dots, x_n\} \) and \( Y = \{y_1, \dots, y_m\} \), DTW computes a warping path \( P = \{(i_k, j_k)\}_{k=1}^L \subseteq [1, n] \times [1, m] \) such that the cumulative cost of alignment is minimized:
\[
\text{DTW}(X, Y) = \min_P \sum_{k=1}^L C(x_{i_k}, y_{j_k}),
\]
where \( C(x_i, y_j) \) is a local distance function between elements, typically the squared Euclidean distance between their embeddings. The path \( P \) must satisfy monotonicity and continuity constraints, ensuring a valid sequential alignment. The cumulative cost matrix \( D \in \mathbb{R}^{n \times m} \) is computed recursively as:
\[
D(i, j) = C(x_i, y_j) + \min \left\{ D(i{-}1, j),\ D(i, j{-}1),\ D(i{-}1, j{-}1) \right\},
\]
with suitable initialization at the borders.

\textbf{Subsequence DTW (S-DTW)} extends DTW to handle cases where one sequence (typically the query) is shorter than the other. Instead of aligning the full sequences, S-DTW searches for a contiguous subsequence of the longer sequence that best aligns with the entire shorter sequence. Formally, given a shorter query sequence \( X \in \mathbb{R}^{n \times d} \) and a longer reference sequence \( Y \in \mathbb{R}^{m \times d} \), where \( n < m \), the S-DTW distance is defined as:
\[
\text{S-DTW}(X, Y) = \min_{1 \leq s \leq e \leq m} \text{DTW}(X, Y_{s:e}),
\]
where \( Y_{s:e} \) denotes the subsequence of \( Y \) from index \( s \) to \( e \). In practice, the cost is computed efficiently by evaluating DTW between \( X \) and all prefixes of \( Y \), and taking the minimum over possible start and end points, as described in~\cite{giorgino2009computing}.

\textbf{Application to Sub-Trajectory Retrieval:}  
In imitation learning, particularly when learning from prior demonstrations, it is often desirable to match short sub-trajectories from a target task to relevant subsequences from longer demonstrations in a prior dataset. S-DTW provides a principled method to do so, as it allows aligning a fixed-length query trajectory to arbitrary contiguous segments in a reference trajectory.

In our work, we first segment each target demonstration into sub-trajectories using a velocity-based heuristic, as described in the main text. For each modality \( f \), we use the associated encoder \( \mathcal{F}_f \) to embed observations into a feature space. Given a target sub-trajectory \( t' = \{o_1, \dots, o_n\} \) and a longer prior trajectory \( t = \{o'_1, \dots, o'_m\} \), we compute a pairwise cost matrix \( C \in \mathbb{R}^{n \times m} \), where:
\[
C(i,j) = \left\| \mathcal{F}_f(o_i) - \mathcal{F}_f(o'_j) \right\|_2^2.
\]
We then apply S-DTW to compute the minimal cumulative alignment cost between the entire query \( t' \) and any contiguous subsequence of \( t \). This produces a similarity score between the target sub-trajectory and the reference. We repeat this process across the prior dataset and retrieve the top-\( K \) most similar subsequences per modality.

For efficient implementation and further details, we refer the reader to standard references on DTW and S-DTW~\cite{giorgino2009computing} as well as the STRAP framework~\cite{memmel2024strap}, which operationalizes this approach for sub-trajectory matching in the context of policy learning.

\end{document}